\documentclass[]{llncs}
\usepackage{amsmath,amssymb}
\usepackage{subfigure}
\usepackage{xspace}

\usepackage{url}
\usepackage{tikz}
\usetikzlibrary{automata,arrows,decorations.pathmorphing,backgrounds,fit,positioning}

\usepackage{color}

\newcommand{\tx}[1]{\text{\em #1}}
\def\Lang{\mathcal{L}}
\newcommand{\tuple}[1]{\langle #1 \rangle \xspace}

\title{Equilibrium Graphs\thanks{This research was partially supported by MINECO project TIN2013-42149-P, Spain.}}

\author{Pedro Cabalar \and Carlos P\'erez \and Gilberto P\'erez}
\institute{Department of Computer Science\\
University of Corunna, Spain\\
\email{\{cabalar,c.pramil,gperez\}@udc.es}
}

\begin{document}

\maketitle
\begin{abstract}
In this paper we present an extension of Peirce's existential graphs to provide a diagrammatic representation of expressions in Quantified Equilibrium Logic (QEL). Using this formalisation, logical connectives are replaced by encircled regions (circles and squares) and quantified variables are represented as ``identity'' lines. Although the expressive power is equivalent to that of QEL, the new representation can be useful for illustrative or educational purposes.
\end{abstract}

\section{Introduction}

Most efforts in Knowledge Representation (KR) have been traditionally focused on symbolic manipulation and, in particular, on logical formulation.
The use of a formal representation is surely convenient for automated reasoning, since computer languages provide nowadays excellent tools for symbolic representation and processing.
Unfortunately, something that is simpler for computational treatment is not always necessarily better for human understanding.
Educational experiences show that learning and understanding logical notation takes some time and effort to novel students.
Even for an experienced student, reading a formula that nests different quantifiers, connectives or parentheses may become a difficult task and lead to errors in formal specification.

One alternative to formal languages that is probably closer to human's intuition is the use of graphical or diagrammatic representations. 
In fact, diagrammatic KR has also been explored and used in different fields of Artificial Intelligence -- a prominent example is, for instance, Sowa's \emph{conceptual graphs}~\cite{Sow76,Sow08}. 
But the use of diagrams for logical representation is older than KR and AI, and actually comes from the very origins of modern philosophical logic.
As commented by Sowa in~\cite{Sow13}, the use of diagrams in logic was something common before the introduction of the current notation, conceived by Peano\footnote{Peano's quantifiers $\exists, \forall$ correspond to the inverted letters E and A, whereas $\vee$ comes from Latin \emph{vel} (``or'') and conjunction $\wedge$ from its inversion.} in 1889~\cite{Pea89}.
In fact, Frege's original formulation of Predicate Calculus already included some diagrammatic component.
But it was Charles Sanders Peirce who first introduced\footnote{Peirce's first proposals of existential graphs date back to 1882, even earlier than Peano's publication of the modern symbolic notation.} a full-blown non-symbolic system for first-order logic: \emph{existential graphs}~\cite{Pei06} (EGs). 
This graphical system allows a complete characterisation of First-Order Logic in diagrammatic terms, without using logical connectives.
However, save few exceptions (like their influence in Sowa's conceptual graphs~\cite{Sow13}), the truth is that EGs did not gain the same popularity as the symbolic notation for classical logic, even though they provide an elegant and simple representation that seems very suitable for educational purposes.
Perhaps one of the difficulties for their consolidation has to do with their strong dependence on classical logic.
Existential graphs take conjunction, negation and existential quantifiers as primitive constructors, building all the rest (disjunction, implication or universal quantification) as derived operations.
This approach leaves no room for other non-classical logics such as intuitionistic or intermediate logics, where we may need keeping all these connectives independently.

In this paper we study an extension of existential graphs to be used as an alternative diagrammatic notation for \emph{Answer Set Programming}~\cite{MT99,Nie99,BET11} (ASP) and, in particular, for its logical formalisation in terms of \emph{Equilibrium Logic}~\cite{Pea96}.
Proposed by David Pearce, Equilibrium Logic has allowed the application of the \emph{stable model} semantics~\cite{GL88}, originally defined for the syntax of logic programs, to the  case of arbitrary propositional formulas.
Moreover, the extension to the first-order case, known as \emph{Quantified Equilibrium Logic}~\cite{PV08}, provides nowadays a general logical notion of stable models for arbitrary theories expressed in the syntax of First-Order Logic.
Equilibrium Logic is defined by imposing a model selection criterion on top of a monotonic intermediate logic known as the logic of \emph{Here-and-There}~\cite{Hey30} (HT).
In this logic, implication is a primitive operation and, although disjunction can be defined in terms of the former plus conjunction, its representation as a derived operator is rather cumbersome.
Something similar happens in \emph{Quantified HT}~\cite{PV08} where, again, the existential quantifier is definable in terms of the universal one, but it is much more convenient to treat both of them as primitive connectives.
In the paper, we extend EGs to allow dealing with all these operators independently by just adding a new graphical primitive (rectangles) to the closed curves and lines already existing in EGs.

This short note constitutes a preliminary proposal, providing several illustrative examples of the potential use, mostly for educational or visualisation purposes.

The rest of the paper is organised as follows.
In the next section, we provide an overview of Existential Graphs, both \emph{alpha} graphs corresponding to propositional logic, and \emph{beta} graphs for first-order logic.
In Section~\ref{sec:qel}, we summarise the main definitions of Quantified Equilibrium Logic, assuming a static Herbrand domain, which is the most common case in ASP.
The main contributions are presented in Sections~\ref{sec:alpha} and \ref{sec:beta} that respectively introduce the extensions of alpha and beta graphs for Equilibrium Logic, providing some examples of their use.
Finally, Section~\ref{sec:conc} concludes the paper.

\section{Existential Graphs}\label{sec:eg}

We recall next the essential components of existential graphs.
Peirce classified EGs into three types, \emph{alpha}, \emph{beta} and \emph{gamma}, that respectively correspond to Propositional Calculus, First-Order Logic with equality and (a kind of) normal modal logic.
We start defining alpha graphs as follows.
A \emph{diagram} in alpha graphs is recursively defined as one of the following:
\begin{itemize}
\item the main page (when empty, it represents truth)
\item atomic propositions
\item a region encircled by a closed curve (called \emph{cut}), which denotes the negation of the subdiagram inside the region. An empty cut represents falsity.
\item finally, although it is not a drawing in itself, the inclusion of several elements inside the same region or cut (including the full page) is implicitly understood as their conjunction
\end{itemize}

\begin{figure}[htbp]
\centering
\subfigure[An example]{\label{fig:rain1}
\includegraphics[scale=1]{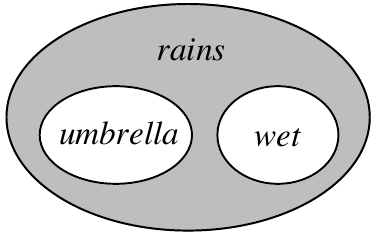}}
\hspace{15pt}
\centering
\subfigure[$p\rightarrow q$]{\label{fig:imp}
\includegraphics[scale=1]{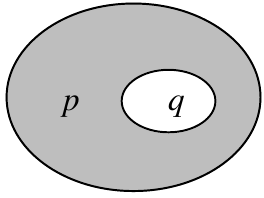}}
\hspace{15pt}
\centering
\subfigure[$p \vee q$]{\label{fig:or}
\includegraphics[scale=1]{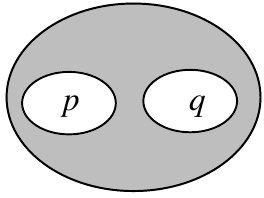}}
\caption{Some alpha graphs.}
\label{fig:alpha}
\end{figure}

As an example, Fig.~\ref{fig:rain1} explicitly represents the formula $\neg (\tx{rains} \wedge \neg \tx{umbrella} \wedge \neg \tx{wet})$ which can also be seen as the implications $\tx{rains} \wedge \neg \tx{umbrella} \to \tx{wet}$ or $\tx{rains} \wedge \neg wet \to \tx{umbrella}$ or the disjunction $\neg \tx{rains} \vee \tx{umbrella} \vee \tx{wet}$, since all these representations are equivalent in classical propositional logic.
Using conjunction and negation as primitive operators, we can easily represent an implication $p \rightarrow q$ as $\neg (p \wedge \neg q)$ (Fig.~\ref{fig:imp}) and a disjunction $p \vee q$ as $\neg(\neg p \wedge \neg q)$ (Fig.~\ref{fig:or}).
Another common feature shown in these examples is that areas encircled by an odd number of cuts (negative areas) are sometimes shaded to facilitate the visualisation.

The alpha system was accompanied by a set of inference and equivalence (diagram redrawing) rules that was proved to be sound and complete with respect to Propositional Calculus (note that, at the time, Tarskian model-based semantics had not been developed yet).
In this preliminary proposal, we will focus on the representation itself, leaving graphical inference in the logic of HT or even in (non-monotonic) Equilibrium Logic for a future study.

\begin{figure}[htbp]
\centering
\subfigure[]{\label{fig:parked}
\includegraphics[scale=1]{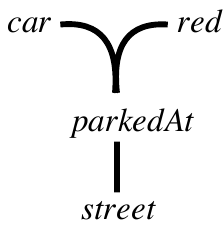}}
\hspace{70pt}
\centering
\subfigure[]{\label{fig:loves}
\includegraphics[scale=1]{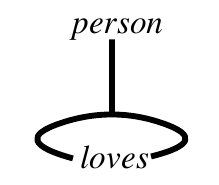}}
\\
\centering
\subfigure[]{\label{fig:mortal}
\includegraphics[scale=1]{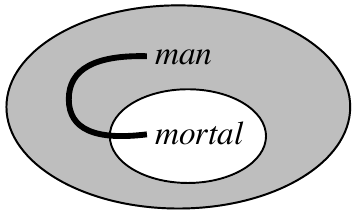}}
\hspace{20pt}
\centering
\subfigure[]{\label{fig:catholic}
\includegraphics[scale=1]{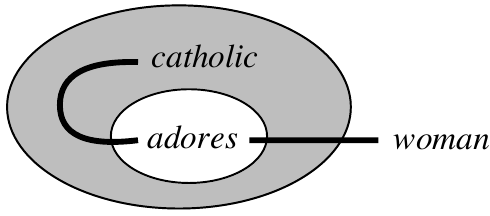}}
\caption{Examples of beta graphs.}
\label{fig:beta}
\end{figure}

For representing first-order expressions, Peirce extended alpha graphs to \emph{beta} graphs by the inclusion of a new type of component in the diagram, \emph{lines of identity}.
A line of identity is an open line that connects one or more atom names.
When it is used to connect more than two atom names, the identity line may bifurcate as many times as needed, getting the shape of a tree or a spider with several ramifications.
The reading for an identity line is an existential quantifier: ``there exists some individual such that \dots''
Figure~\ref{fig:beta} shows several examples.
Fig.~\ref{fig:parked} asserts that there is a red car parked at a street: $\exists x \exists y (car(x) \wedge red(x) \wedge parkedAt(x,y) \wedge street(y))$.
Fig.~\ref{fig:loves} means that there is some person that loves herself, $\exists x (person(x) \wedge loves(x,x))$.
Fig.~\ref{fig:mortal} says that every man is mortal, $\neg \exists x (man(x) \wedge \neg mortal(x))$ or, if preferred, $\forall x (man(x) \to mortal(x))$.
Finally, Fig.~\ref{fig:catholic} specifies that there is a woman adored by every catholic: $\exists x ( woman(x) \wedge \forall y (catholic(y) \to adores(y,x)))$.

As we can see, identity lines introduce a subtle difference in the role of atom names in beta graphs.
Atoms represent now $n$-ary predicates whose arguments correspond to imaginary place holders surrounding the atom name that are used as endpoints of identity lines.
In the case of unary predicates, such as $man$ or $car$, the position of this place holder is irrelevant.
However, when the predicate arity is greater than one, the argument location becomes relevant: for instance, in Fig.~\ref{fig:beta}, predicate $adores$ has a left argument that corresponds to the adorer and a right argument corresponding to the adored person.

Another important observation is that beta graphs do not provide a specific method for representing constants.
For instance, there is no way for expressing that every catholic adores (Virgin) Mary other than using a unary predicate $\tx{Mary}$ to designate that specific person instead of some abstract $\tx{woman}$.

One final remark on identity lines is that they can be actually seen as an implicit equality predicate.
Some representations even introduce a label ``is'' for the identity line to emphasize this feature.
Following this interpretation, when an identity runs through an empty cut we get a convenient way to represent an inequality of the form $x \neq y$.
As an example, the diagram in Figure~\ref{fig:god} represents the monotheist sentence ``there is a God and only one God.''
\begin{center}

\begin{figure}[htbp]
\centering
\includegraphics{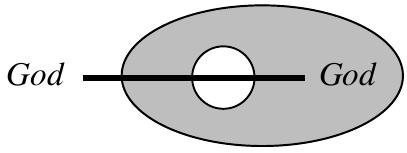} 
\caption{$\exists x (God(x) \wedge \neg \exists y (x \neq y \wedge God(y)))$}
\label{fig:beta}
\end{figure}

\end{center}

\section{Quantified Equilibrium Logic}\label{sec:qel}

For the sake of completeness, we recall in this section the basic definitions of Quantified Equilibrium Logic for function-free theories and Herbrand domains, since this is the most frequent situation in ASP.
We consider first-order languages $\mathcal{L}=\tuple{D,P}$ built  over a set of \emph{constant} symbols, $D$ (the Herbrand domain), and a set of \emph{predicate} symbols, $P$. 
The sets of $\mathcal{L}$-formulas, $\mathcal{L}$-sentences and atomic $\mathcal{L}$-sentences are defined in the usual way.
If $D$ is a non-empty set, we denote by $At(D,P)$ the set of ground atomic sentences of the language $\langle D, P\rangle $.
By an $\mathcal{L}$-interpretation $I$ over a set $D$ we mean a subset of $At(D,P)$. 
A \emph{classical} Herbrand $\Lang$-structure can be regarded as a tuple $\mathcal{M} = \langle D, I\rangle$ where $I$ is an $\mathcal{L}$-interpretation over $D$.

A \emph{here-and-there} ${\cal L}$-\emph{structure} is a tuple $\mathcal{M} = \langle D, I_h, I_t\rangle$ where $\langle D,  I_h\rangle$ and $\langle D, I_t\rangle$ are classical Herbrand $\Lang$-structures such that $I_h\subseteq I_t$.
We say that the structure is \emph{total} when $I_h=I_t$.
We can think of a here-and-there structure $\mathcal{M}$ as similar to a first-order classical model, but having two parts, or components, $h$ and $t$, that correspond to two different points or ``worlds'', `here' and  `there',  in the sense of Kripke semantics for intuitionistic logic, where the worlds are ordered by $h \leq t$. 

We assume that $\mathcal{L}$ contains the constants $\top$ and $\bot$ and regard $\neg \varphi$ as an abbreviation for $\varphi \rightarrow \bot$.
Satisfaction of formulas is defined as follows.
Given some world $w \in \{h,t\}$:

\begin{itemize}
\item
$\mathcal{M},w\models \top ,  \; \mathcal{M},w \not\models \bot$
\item $\mathcal{M},w\models p$ iff $p \in I_w$ for any atom $p \in At(D,P)$
\item $\mathcal{M},w\models c=d$ iff $c$ and $d$ denote the same constant from $D$
\item $\mathcal{M},w\models \varphi \land \psi$ iff $\mathcal{M},w\models
\varphi$ and $\mathcal{M},w\models \psi$.
\item
$\mathcal{M},w\models \varphi \lor \psi$ iff $\mathcal{M},w\models
\varphi$ or $\mathcal{M},w\models \psi$.
\item
$\mathcal{M},t\models \varphi \to \psi$ iff $\mathcal{M},t
\not\models \varphi$ or $\mathcal{M},t \models \psi$.
\item
$\mathcal{M},h\models \varphi \to \psi$ iff $\mathcal{M},t\models
\varphi \to \psi$ and $\mathcal{M},h \not\models \varphi$ or
$\mathcal{M},h \models \psi$.
\item
$\mathcal{M},w\models \forall x \varphi(x)$ iff
$\mathcal{M},w\models\varphi(d)$ for all $d\in D$.
\item
$\mathcal{M},w\models \exists x \varphi(x)$ iff
$\mathcal{M},w\models\varphi(d)$ for some $d\in D$.
\end{itemize}

We say that $\mathcal{M}$ is a \emph{model} of a sentence $\varphi$ iff $\mathcal{M},h\models \varphi$.
The resulting logic is called \emph{Quantified Here-and-There Logic with static domains and decidable equality} (QHT, for short).
\begin{definition}[Equilibrium model]
Let $\varphi$ be an $\mathcal{L}$-sentence. An {\em equilibrium} model of $\varphi$ is a total model $\mathcal{M}=\langle
D,I_t,I_t\rangle$ of $\varphi$ such that there is no model of $\varphi$ of the form $\langle
D,I_h,I_t\rangle$ where $I_h$ is a  proper subset of $I_t$. 
\end{definition}

When $\tuple{D,I_t,I_t}$ is an equilibrium model of $\varphi$ we say that the classical (Herbrand) interpretation $\tuple{D,I_t}$ is a \emph{stable model} of $\varphi$.

\section{Equilibrium Alpha Graphs}\label{sec:alpha}

\begin{figure}[htbp]
\centering
\subfigure[$\tx{toss} \to \tx{head} \vee \tx{tails}$]{
\label{fig:toss}
\centering
\includegraphics{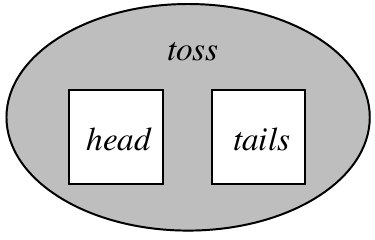}
}
\hspace{1cm}
\subfigure[$\tx{rains} \wedge \neg \tx{umbrella} \to \tx{wet}$]{
\label{fig:rain2}
\centering
\includegraphics{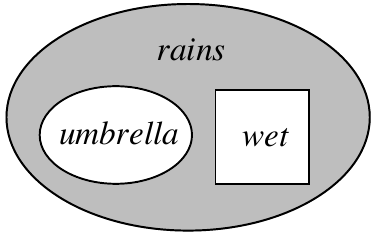}
}
\hspace{1cm}
\subfigure[$\tx{red}\vee\tx{orange}\vee\tx{green}$]{
\label{fig:or2}
\centering
\includegraphics{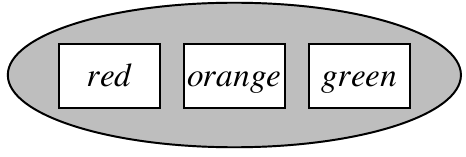}
}\caption{Examples of conditionals.}
\label{fig:basic}
\end{figure}

Let us begin considering the use of alpha graphs to represent Equilibrium Logic theories (or ASP logic programs).
A first difficulty we face is that implication is a primitive operator in HT, and cannot be represented in terms of conjunction and disjunction (see Theorem~4 in~\cite{ACP+15}).
This generates a conflict with the use of material implication in alpha graphs, defined in terms of negation and conjunction.
To overcome this problem, we replace the cut component (negation) by a new diagrammatic construction we will simply call \emph{conditional}.
A conditional has the form of a closed curve (or ellipse) and may contain inside a number $n \geq 0$ of rectangles we call \emph{consequents}.
Intuitively, when all the elements inside the ellipse (but not in the rectangles) hold then one of the rectangles must hold (that is, we implicitly have a disjunction of consequents).
As an example, Fig.~\ref{fig:toss} represents the implication $toss \to head \vee tails$.
The case of $0$ rectangles corresponds to an implication with $\bot$ (the empty disjunction) as a consequent.
In other words, a conditional without consequents is just read as a negation, as happens in Peirce's alpha diagrams.
As an example, Fig.~\ref{fig:rain2} represents now the implication $\tx{rains} \wedge \neg \tx{umbrella} \to \tx{wet}$, that is, $\tx{rains} \wedge (\tx{umbrella} \to \bot) \to \tx{wet}$.
It is perhaps worth to compare to Fig.~\ref{fig:rain1} where, as we commented before, there was no way to differentiate between a negative condition in the antecedent and a positive condition in the consequent ($\tx{wet}$ and $\tx{umbrella}$ played the same role).
This reflected the non-directional nature of material implication.
Under our new notation, Fig.~\ref{fig:rain2} allows now distinguishing the elements in the consequent ($\tx{wet}$ is inside a rectangle) from those in the antecedent, either negated ($\tx{umbrella}$) or not ($\tx{rains}$).

As we have seen, when the conditional has no consequents, it corresponds to a negation. 
In an analogous way, when the conditional has an an empty antecedent (it only contains rectangles) it obviously represents a disjunction.
Fig.~\ref{fig:or2} represents the disjunction $\tx{red}\vee\tx{orange}\vee\tx{green}$ for the possible colors of a traffic light.

\begin{figure}[htbp]
\centering
\includegraphics{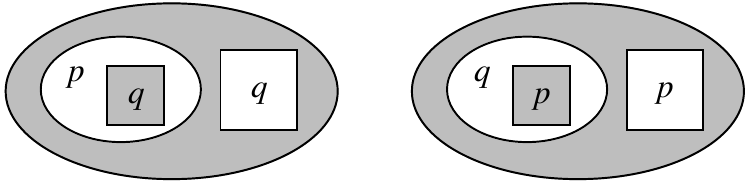}
\caption{$((p \to q) \to q) \wedge ((q \to p)\to p)$}
\label{fig:or3}
\end{figure}

A disjunction $p \vee q$ in HT can be defined in terms of conjunction and implication, as it is equivalent to the expression 
$$((p \to q) \to q) \wedge ((q \to p)\to p)$$
whose diagrammatic representation is shown in Figure~\ref{fig:or3}.
However, the only ``advantage'' we would gain using this representation (as primitive for disjunction) is that we would not need more than one rectangle in each conditional, while we would clearly lose readability.

\begin{figure}[htbp]
\centering
\subfigure[$(\neg q \to p) \wedge (\neg p \to q)$]{
\label{fig:choice1}
\centering
\includegraphics{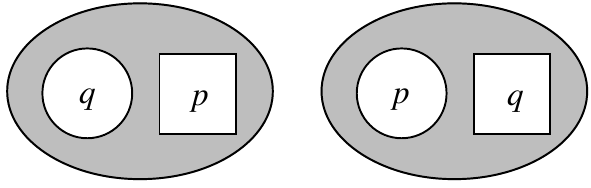}
}
\hspace{30pt}
\subfigure[$p \vee \neg p$]{
\label{fig:choice2}
\centering
\includegraphics{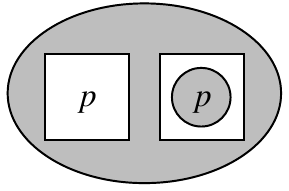}
}

\caption{Choice rules.}
\label{fig:choice}
\end{figure}

An interesting construction in ASP is the use of choice rules.
The original way to build a choice that causes the non-deterministic addition of an atom $p$ in ASP was using some auxiliary predicate $q$ and building an even negative cycle as the one shown in Figure~\ref{fig:choice1}.
A second possibility that does not require an auxiliary predicate is using the formula $p\vee \neg p$ (which is not a tautology in HT) represented in Figure~\ref{fig:choice2}.

\begin{figure}[htbp]
\centering
\subfigure[$\tx{bird} \wedge \neg \tx{abnormal} \to \tx{flies}$]{
\label{fig:penguin1}
\centering
\includegraphics{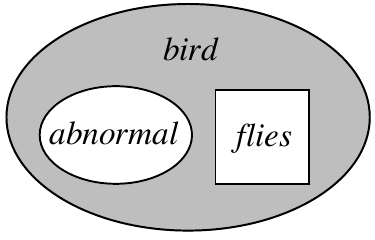}
}
\hspace{30pt}
\subfigure[$\tx{penguin} \! \to \!\! \tx{abnormal} \wedge\!\tx{bird}$]{
\label{fig:penguin2}
\centering
\includegraphics{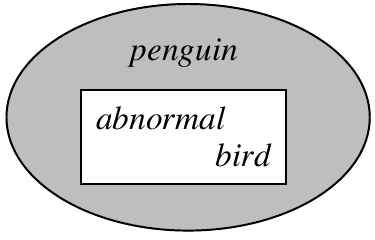}
}

\caption{Birds and penguins.}
\label{fig:penguin}
\end{figure}

To conclude this section, we illustrate a typical example from Non-Monotonic Reasoning.
Fig.~\ref{fig:penguin} encodes a propositional program with two rules respectively asserting that a bird normally flies and that a penguin is an abnormal bird.
 
\section{Equilibrium Beta Graphs}\label{sec:beta}

\begin{figure}[htbp]
\centering
\subfigure[$\forall x (\tx{man}(x) \to \tx{mortal}(x))$]{
\label{fig:forall}
\centering
\includegraphics{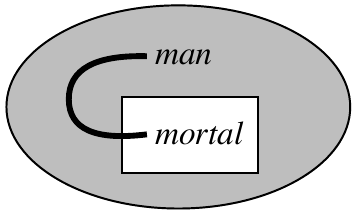} \hspace{10pt}
}
\hspace{1cm}
\subfigure[$\exists x (\tx{man}(x) \to \tx{mortal}(x))$]{
\label{fig:forall2}
\centering
\includegraphics{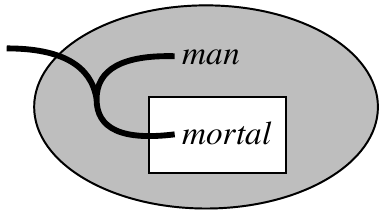}
}
\hspace{1cm}
\subfigure[$\forall x(\tx{man}(x) \to \tx{mortal})$]{
\label{fig:forall3}
\centering
\includegraphics{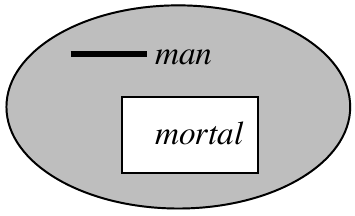}
}\caption{Some conditionals with identity lines.}
\label{fig:idcond}
\end{figure}

As happened with implication in the propositional case, the universal quantifier is a primitive operator in QEL and cannot be represented in terms of existential quantifiers and the other connectives\footnote{It is actually the other way around.
Any existentially quantified formula $\exists x P(x)$ is QHT equivalent to $\forall x \forall y ((P(x) \to P(y)) \to P(y))$.}.
Therefore, introducing identity lines would not suffice to cover the expressive power of QEL if they were always read as existential quantifiers.
Fortunately, since we count with a new conditional connective, whose expressiveness is richer than the simple cut, we can use it to cover both existential and universal quantifiers as follows.
Any identity line encircled by a conditional and with some portion outside any consequent (rectangle) of that conditional corresponds to a universal quantifier.
Figure~\ref{fig:idcond} shows some examples combining conditionals and identity lines.
Fig.~\ref{fig:forall} corresponds to a universal quantifier, saying that all men are mortal:
\begin{eqnarray}
\forall x (\tx{man}(x) \to \tx{mortal}(x)) \label{fig:allmen}
\end{eqnarray}
Note the difference with respect to the version in Fig.~\ref{fig:mortal}, where we had a cut (negation) instead of the rectangle.
This version is still a correct equilibrium beta graph, but its reading corresponds to:
\begin{eqnarray}
\forall x (\tx{man}(x) \wedge \neg \tx{mortal}(x) \to \bot) \label{fig:nomen} 
\end{eqnarray}
which has a quite different meaning from \eqref{fig:allmen}: the latter is a rule that allows deriving $\tx{mortal}(x)$ from any $\tx{man}(x)$, whereas \eqref{fig:nomen} acts as a constraint, forbidding stable models where some man is not known to be mortal.
Another equivalent reading of a constraint like Fig.~\ref{fig:mortal} (i.e. a conditional with existential lines but no consequents) is just as a negation of an existential quantifier:
\begin{eqnarray*}
\neg \exists x (\tx{man}(x) \wedge \neg \tx{mortal}(x))
\end{eqnarray*}

Fig.~\ref{fig:forall2} corresponds to an existential quantifier: it contains an identity line which is not encircled by the conditional (it ``comes from outside'').
As for Fig.~\ref{fig:forall3}, it represents a universal quantifier: it is encircled by the conditional and (fully) outside the consequent, representing the formula $\forall x(\tx{man}(x) \to \tx{mortal})$.
However, a case like this (the line does not cross any rectangle) can also be read as an existential quantifier in the antecedent, since the last formula is equivalent to $( \exists x \ \tx{man}(x) ) \to \tx{mortal}$.

To conclude this section, we provide an example encoding the well-known Hamiltonian cycle problem: given a graph $G$, find a cyclic path that visits each node in $G$ exactly once.
We assume that the graph $G$ is provided in terms of facts for the binary predicate $\tx{edge}$, related to node names.
The Hamiltonian path is encoded using a binary predicate $in$, meaning that the corresponding edge is included in the path, for a given stable model.

\begin{figure}[htbp]
\centering
\includegraphics[scale=1]{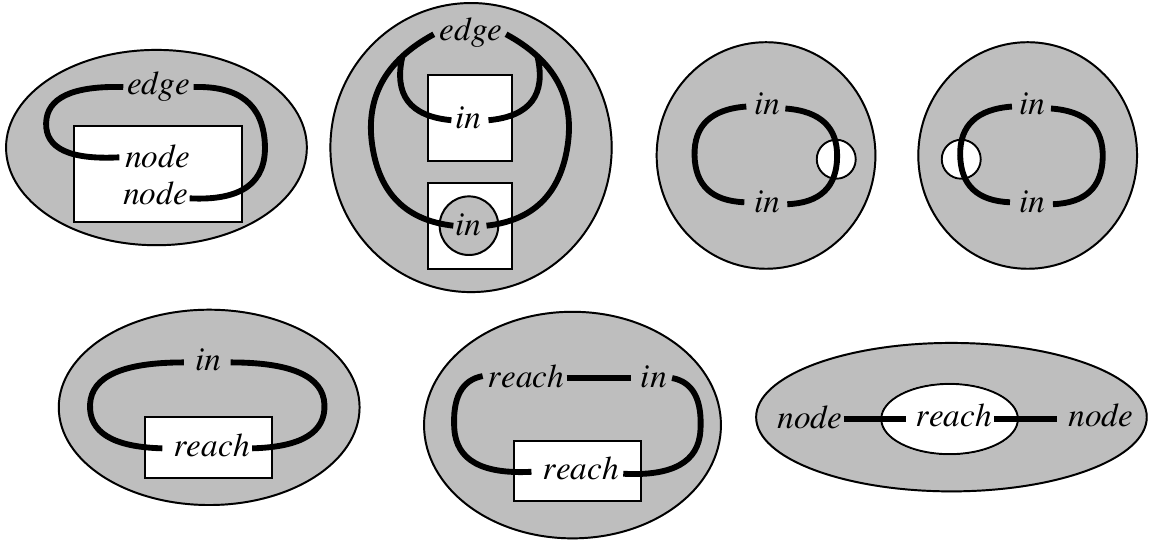}
\caption{Graphical encoding of the Hamiltonian cycles problem.}
\label{fig:hamilt}
\end{figure}

Figure~\ref{fig:hamilt} shows a possible diagrammatic representation of this problem.
The corresponding formulas, reading all the conditionals from left to right and from up to down would respectively be:
\begin{eqnarray}
\forall x \forall y & & \big(\ edge(x,y) \to node(x) \wedge node(y)\ \big) \label{f:h1}\\
\forall x \forall y & & \big(\ edge(x,y) \to in(x,y) \vee \neg in(x,y)\ \big) \label{f:h2}\\
\neg \exists x \exists y \exists z & & \big( in(x,y) \wedge in(x,z) \wedge y \neq z\ \big) \label{f:h3}\\
\neg \exists x \exists y \exists z & & \big( x \neq y \wedge in(x,z) \wedge in(y,z)\ \big) \label{f:h4}\\
\forall x \forall y & & \big(\ in(x,y) \to reach(x,y)\ \big) \label{f:h5}\\
\forall x \forall y \forall z& & \big(\ reach(x,y) \wedge in(y,z) \to reach(x,z)\ \big) \label{f:h6}\\
\neg \exists x \exists y & & \big( node(x) \wedge node(y) \wedge \neg reach(x,y)\ \big) \label{f:h7}
\end{eqnarray}
All of them can be easily rewritten as program rules in standard ASP syntax.
\eqref{f:h1} asserts that the two arguments of predicate $edge$ are nodes.
\eqref{f:h2} is a non-deterministic choice to include any edge in the stable model or not.
\eqref{f:h3} and \eqref{f:h4} are constraints respectively forbidding that two edges in the path with the same origin go to two different targets, and vice versa, that two different origin nodes go to a common target.
Formulas \eqref{f:h5},\eqref{f:h6} define the transitive closure $reach$ of predicate $in$.
Finally, \eqref{f:h6} is a constraint forbidding that a node $y$ cannot be reached from another node $x$.

\begin{figure}[htbp]
\centering
\includegraphics[scale=0.9]{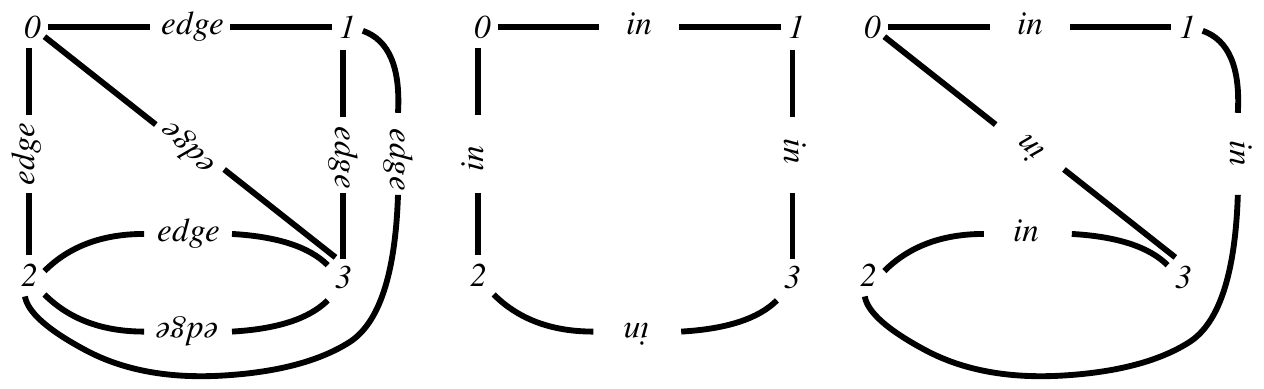}
\caption{A graph and its two stable models corresponding to the Hamiltonian paths.}
\label{fig:hamilt2}
\end{figure}

Figure~\ref{fig:hamilt2} shows three diagrams respectively depicting a possible example of input graph (facts for predicate $edge$) plus the two corresponding stable models that represent the Hamiltonian paths of the input graph.

Given an Equilibrium Beta Graph $G$ we define its corresponding Peirce's Beta Graph $G^*$ as the result of replacing all rectangles in $G$ by ellipses\footnote{We could alternatively say that $G^*$ is just ``Peirce's reading'' of $G$ without need of any transformation, since a rectangle is also a case of closed curve, and Peirce's original approach would make no real distinction between ellipses and rectangles.}.

\begin{proposition}
Let $\Phi(G)$ denote the first-order formula associated to $G$ under the interpretation in the current paper and $\Phi(G^*)$ the formula associated to $G^*$ under Peirce's beta graphs interpretation. Then $\Phi(G)$ and $\Phi(G^*)$ are equivalent in classical First-Order Logic.\qed
\end{proposition}

\section{Conclusions}\label{sec:conc}

By introducing a minimal variation on Peirce's existential graphs (the introduction of rectangles), we have presented a diagrammatic representation of Quantified Equilibrium Logic and ASP programs.
In fact, the current formulation allows representing any intermediate logic, since it has just allowed defining implication, disjunction and universal quantification as primitive constructions, rather than derived operators in terms of conjunction, negation and existential quantifiers.

This note constitutes a preliminary proposal but, obviously, much work is left to do yet.
First, it is unclear yet how to provide a fully visual semantic characterisation, especially for beta diagrams.
Another desirable feature would be a set of inference and equivalence diagram-rewriting rules that covered QHT in a sound and complete way.

Under the ASP perspective, it is interesting to note that not every QEL formula has a direct correspondence to an ASP program: for instance, formulas beginning by existential quantifiers are not ASP representable.
It would also be interesting to identify graphical features about the kind of diagrams that have a direct translation into ASP.
This also includes the graphical characterisation of \emph{safety} conditions, required for a suitable grounding.
The introduction of complex ASP constructs such as aggregates or preferences, or even the diagrammatic representation for arithmetic expressions constitutes and important difficulty to be solved yet.
Finally, regarding implementation, a promising line to explore would be the integration of this graphical notation into a visual tool for ASP like the one in~\cite{FRR10}.

\bibliographystyle{splncs}
\bibliography{refs}
\end{document}